\documentclass{article}

     \PassOptionsToPackage{numbers, compress}{natbib}

\usepackage[final]{neurips_2020}

\usepackage[utf8]{inputenc} 
\usepackage[T1]{fontenc}    
\usepackage{hyperref}       
\usepackage{url}            
\usepackage{booktabs}       
\usepackage{amsfonts}       
\usepackage{nicefrac}       
\usepackage{microtype}      

\usepackage{amsmath,amsfonts,bm}

\def\eqref#1{equation~\ref{#1}}

\def\1{\bm{1}}

\DeclareMathAlphabet{\mathsfit}{\encodingdefault}{\sfdefault}{m}{sl}
\SetMathAlphabet{\mathsfit}{bold}{\encodingdefault}{\sfdefault}{bx}{n}

\usepackage{hyperref}
\usepackage{url}
\usepackage{graphicx}
\usepackage{amsmath}
\usepackage{amssymb}
\usepackage{stmaryrd}
\usepackage{multirow}
\usepackage{subcaption}
\usepackage{float}
\usepackage{wrapfig}

\usepackage{subcaption}
\usepackage{graphicx}
\usepackage{lipsum}

\usepackage[font={footnotesize}]{caption}

\title{Towards Good Practices in Self-supervised Representation Learning}

\author{%
  Srikar Appalaraju, Yi Zhu, Yusheng Xie, Istv{\'{a}}n Feh{\'{e}}rv{\'{a}}ri \\
  Amazon Inc. \\
  \texttt{ \{srikara, yzaws, yushx, istvanfe\} @ amazon.com} \\
}

\begin{document}

\maketitle

\begin{abstract}
	Self-supervised representation learning has seen remarkable progress in the last few years. 
	More recently, contrastive instance learning has shown impressive results compared to its supervised learning counterparts.
	However, even with the ever increased interest in contrastive instance learning, it is still largely unclear why these methods work so well.
	In this paper, we aim to unravel some of the mysteries behind their success, which are the \textit{good practices}.
	Through an extensive empirical analysis, we hope to not only provide insights but also lay out a set of best practices that led to the success of recent work in self-supervised representation learning.
\end{abstract}

\setlength{\textfloatsep}{7pt plus 1.0pt minus 2.0pt}
\setlength{\floatsep}{7pt plus 1.0pt minus 2.0pt}
\setlength{\intextsep}{7pt plus 1.0pt minus 2.0pt}

\vspace{-2ex}
\section{Introduction}
\label{sec:introduction}
\vspace{-1ex}


Self-supervised representation learning (SSL) has been a hot area of research in the last several years \cite{gidaris2018unsupervised,noroozi2017representation,doersch2015unsupervised,pathak2016context,noroozi2016unsupervised,zhang2016colorful,caron2018deep,donahue2016adversarial,agrawal2015learning,grill_arxiv2020_byol,caron2020unsupervised}. 
The allure of SSL is the promise of annotation-free ground-truth to ultimately learn a superior data representation (when compared to supervised learning).
More recently, a type of contrastive learning method based on instance discrimination as pretext task~\cite{dosovitskiy_nips2014_exemplar,wu_cvpr2018_instdisc} has taken-off as it has been consistently demonstrated to outperform its supervised counterparts on downstream tasks like image classification and object detection~\cite{oord2018representation,tian2019contrastive,misra2020self,he_cvpr2020_moco,khosla_arxiv2020_supcontrast,chen_icml2020_simclr,caron2020unsupervised}.

Putting the progress in contrastive instance learning aside for a moment, many recent observations seem to contradict what we have known from supervised learning.
For example,~\cite{bachman_nips2019_amdim,chen_icml2020_simclr} have shown that simply adding a nonlinear projection head, i.e., one fully-connected (fc) layer and one activation layer, can significantly improve the quality of learned representations.
Quantitatively, the nonlinear projection head can help to improve top-1 classification accuracy on ImageNet by over $10\%$ in~\cite{chen_icml2020_simclr} and $5.6\%$ in~\cite{chen_arxiv2020_mocov2}.
However, adding such a shallow multilayer perceptron (MLP) head is often not effective in supervised learning.
Take another example, recent methods \cite{he_cvpr2020_moco,chen_icml2020_simclr} adopt aggressive and strong data augmentation during contrastive pre-training. 
Although data augmentation has been proven to be useful, overly aggressive augmentations often lead to worse results in supervised or other self-supervised learning methods.
Then why contrastive instance learning does not suffer from strong data augmentation?
At this moment, there is no concrete evidence to answer these questions.



After closely observing recent contrastive instance learning work, it becomes apparent that what seem to be design choices are in-fact good practices which are largely responsible to their disruptive success. 
Furthermore, some of these good practices can effectively be transferred to other non-contrastive, unsupervised learning methods~\citep{grill_arxiv2020_byol}.
Hence in this work, we focus on the importance of these design choices using extensive experimental evidence.
We hope to provide insights to the self-supervised learning community, with the potential impact and application even beyond it. 
Specifically, our contributions include:  1) We empirically show why MLP head helps contrastive instance learning and visualize it using a feature inversion approach. 2) We present the semantic label shift problem caused by strong data augmentation in supervised learning and study the difference between supervised and contrastive learning. 3) We investigate on negative samples and find that good practices can help to eliminate the need of using large number of them, thereby could simplify the framework design.



\section{Nonlinear Projection Head}
\label{sec:mlp}

\begin{figure}
	\centering
	\begin{minipage}{0.45\textwidth}
		\centering
		\includegraphics[width=7cm,height=3cm]{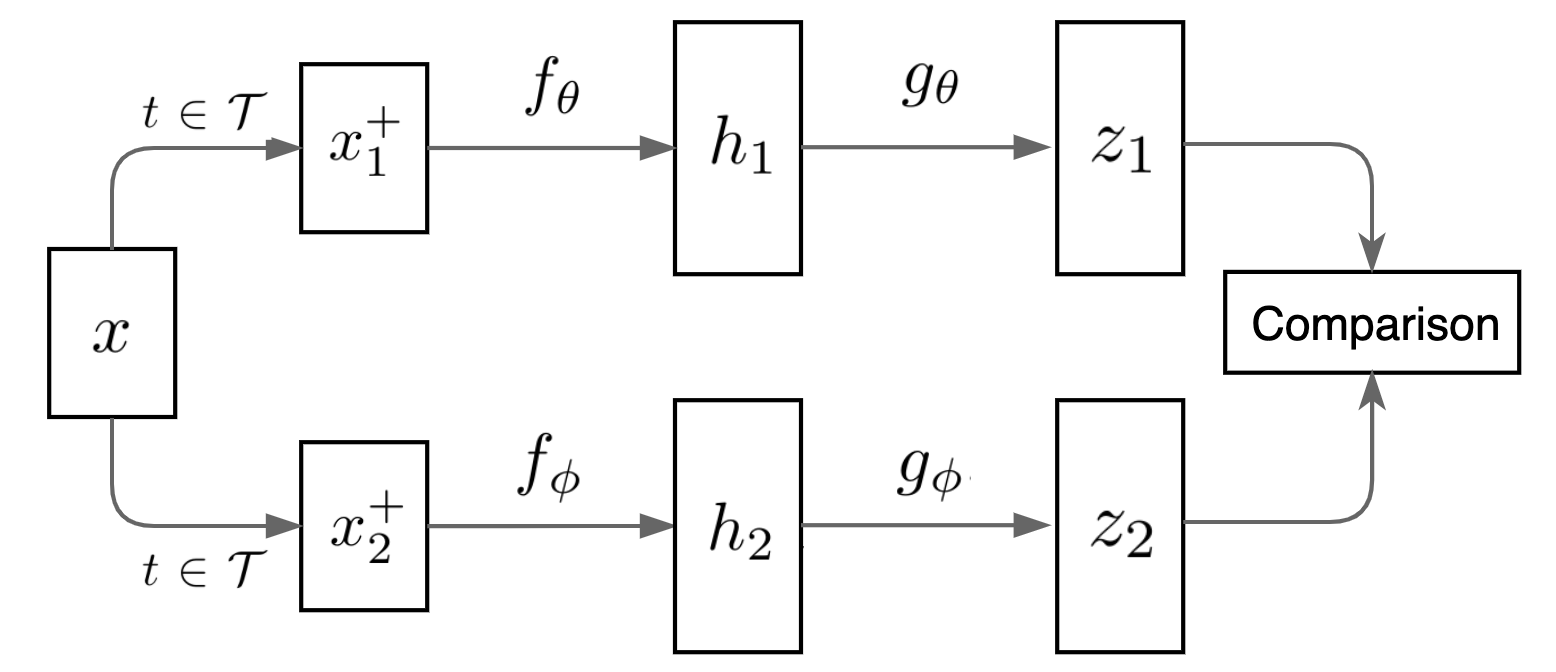}
		\caption{A generic visual depiction of recent contrastive instance learning methods.}\label{fig:splash_instdisc}
	\end{minipage}
	\hspace{6ex}
	\begin{minipage}[0.1\textheight]{0.45\textwidth}
		\centering
		\captionsetup{type=table} 
		\begin{tabular}{lcl}
			\noalign{\smallskip}
			\hline
			\noalign{\smallskip}
			Method         & Top-1 Acc ($\%$)  \\
			\hline
			\noalign{\smallskip}
			MoCov2    &  $64.4$ \\
			(a) no MLP head    & $59.2$ \\
			(b) fixed MLP head    & $62.9$ \\
			(c) deeper MLP head    & $63.9$ \\
			(d) narrower MLP head    & $62.2$ \\
		\end{tabular}
		\caption{Investigation of nonlinear projection head in MoCov2 \cite{chen_arxiv2020_mocov2} on ImageNet.}\label{tab:mlp}
	\end{minipage}
\end{figure}

\begin{figure}[t]
	\begin{center}
		\includegraphics[width=1.0\linewidth,height=3.5cm]{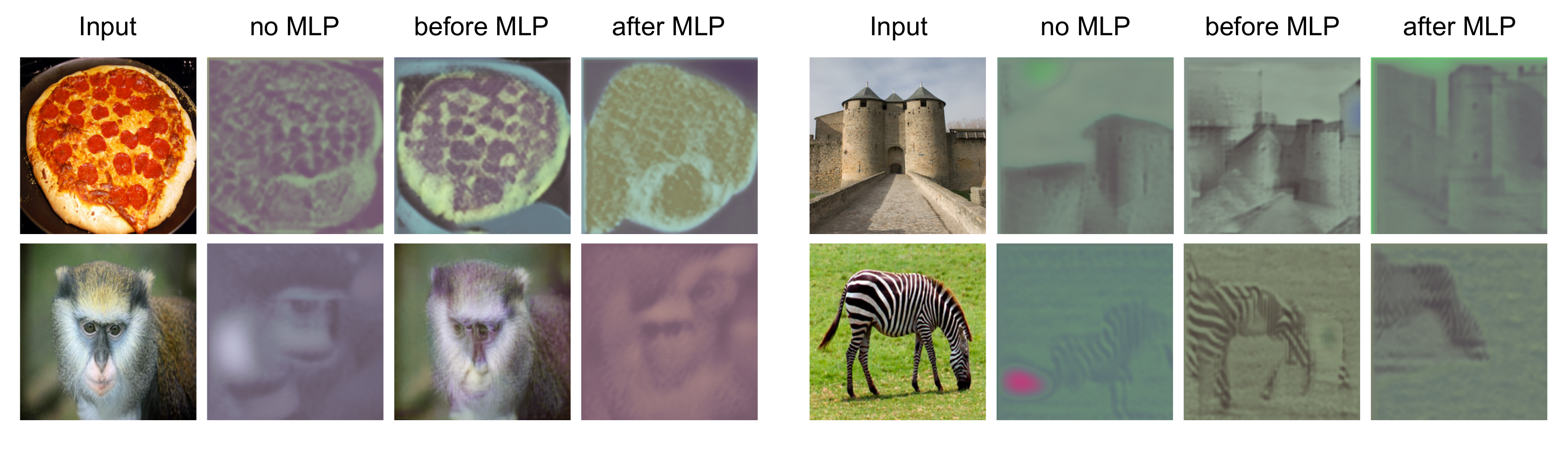}
	\end{center}
	\vspace{-4ex}
	\caption{Visualization of feature inversion results by DIP \citep{ulyanov_cvpr2018_dip}. Each example contains: (a) original image, (b) reconstructed image using $h$ from a model trained without MLP head, (c) using $h$ from a model trained with MLP head and (d) using $z$ from a model trained with MLP head.}
	\label{fig:mlp_dip}
\end{figure}

Before diving into the details, we first revisit recent contrastive instance learning methods and present a generic visual depiction of its framework in Figure \ref{fig:splash_instdisc}. More formally, given a set of images $\mathcal{X}$, an image $x$ is uniformly sampled from $\mathcal{X}$ and is augmented using various augmentation techniques $t \in \mathcal{T}$ to generate the positive pairs $x_1^{+}$ and $x_2^{+}$. $f_{\theta}$ and $f_{\phi}$ are encoders which map the images to visual representations $h_{1}$ and $h_{2}$. These visual representations are then projected via $g_{\theta}$ and $g_{\phi}$, which are often MLP heads, to lower-dimensional features $z_{1}$ and $z_{2}$ for similarity comparison. 
By optimizing InfoNCE loss \cite{oord2018representation}, the model learns to map similar instances closer and dissimilar instances farther apart in the embedding space. 

A main goal of unsupervised representation learning is to learn features that are transferable to downstream tasks. 
Typically the outputs of the penultimate layer $h$ are considered for transferring to other tasks. 
Recently, ~\cite{bachman_nips2019_amdim,chen_icml2020_simclr,chen_arxiv2020_mocov2} have shown that simply adding a MLP head as shown in Figure~\ref{fig:splash_instdisc} can significantly improve the quality of learned feature representation.
However, adding such a shallow MLP head is often not so effective in supervised learning regime.
So the question arises, \textit{why adding a nonlinear projection head is so important for contrastive instance learning}? 

In this paper we attempt to answer this question by designing experiments to explore different aspects of the nonlinear projection head in MoCov2 during unsupervised pre-training. 
First, as shown in Table~\ref{tab:mlp} (a), removing the projection head $g$ significantly degrades the classification accuracy compared to baseline. 
Next, we initialize the nonlinear projection head $g$ with a uniform distribution and freeze its parameters during the unsupervised pre-training. Interestingly, as we can see in Table~\ref{tab:mlp} (b), we obtain better representations compared to removing the projection head. 
This indicates that the nonlinear projection head is useful beyond its learning capability offered by the extra two layers. In fact, it is the nonlinear transformation itself that somehow benefits the learning process even if the parameters of this transformation is randomly initialized.
To strengthen our observation, we investigate two other model variations. 
We first deepen the nonlinear projection head with more hidden layers, i.e., fc $\shortrightarrow$ ReLU $\shortrightarrow$ fc $\shortrightarrow$ ReLU which in theory adds more learning capacity.
As shown in Table~\ref{tab:mlp} (c), this seems not to bring extra benefits. 
We then narrow the nonlinear projection head by reducing the embedding dimensionality, e.g, $2048 \shortrightarrow 128$ for a ResNet50.
As shown in Table~\ref{tab:mlp} (d), such a drastic dimension reduction indeed lowers the performance compared to baseline, but still outperforms setting (a) by a large margin.
With these insights we are more confident to conclude: it is the transformation of the projection head, which separates the pooled convolutional features from the final classification layer, that helps the representation learning. 
But why is such separation beneficial?

We argue that the nonlinear projection head acts as a filter separating the information-rich features useful for downstream tasks (i.e. color, rotation, or shape of objects) from the more discriminative features that are more useful for the contrastive loss.
This conjecture was previously introduced in \cite{chen_icml2020_simclr} and verified by using features to predict transformations applied during the pre-training. 
In this work, we provide further visual evidence to support this hypothesis.

Inspired by deep image prior (DIP) \cite{ulyanov_cvpr2018_dip}, we perform feature inversion to obtain natural pre-images. 
By looking at the natural pre-image, we can diagnose which information is lost and which invariances are gained by the network. 
Specifically, we invert the features before and after the nonlinear projection head, $h$ and $z$ respectively.
As we can see in Figure \ref{fig:mlp_dip}, using features before MLP projection head gives the best reconstruction result. 
Even though we use globally pooled features without spatial dimension, they are able to generate decent image reconstructions, maintaining most color, shape, location and orientation information. 
However, features learned without projection or features after projection head only preserve the most discriminative information to make classification.
This observation supports our claim that layers close to loss computation will lose information due to invariances to data transformations induced by the contrastive loss.


\begin{wrapfigure}{r}{0.45\textwidth}
	\vspace*{-0.090\textwidth}
	\centering 
	\begin{tabular}{ccc}
		\hspace*{-0.025\textwidth}
		\includegraphics[width=0.15\textwidth]{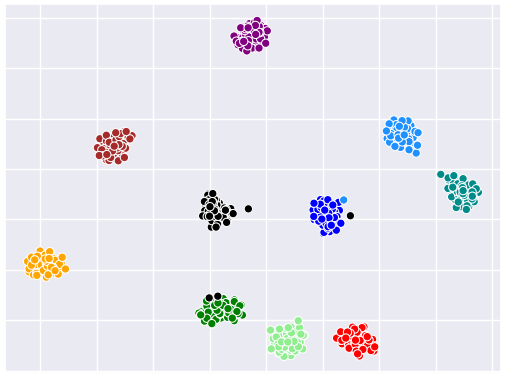}&
		\hspace*{-0.035\textwidth}
		\includegraphics[width=0.15\textwidth]{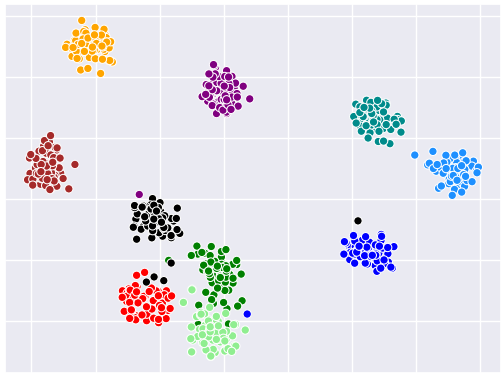}&
		\hspace*{-0.035\textwidth}
		\includegraphics[width=0.15\textwidth]{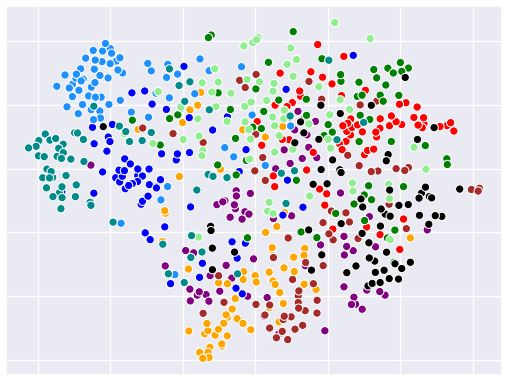}\\
		(a)&(b)&(c)\\
	\end{tabular}
   \vspace*{-0.020\textwidth}
	\caption{$t$-SNE plots from 10 randomly chosen ImageNet classes. We use the 1000-d features from a ImageNet-trained ResNet50. The sub-figures use the same images but under: (a) no; (b) weak;  (c) strong augmentation.}
	\label{fig:tsne}
\end{wrapfigure}

\section{Strong Data Augmentation}
\label{sec:data_aug}

Data augmentation is an important regularization technique in training most deep learning models, ranging from AlexNet~\cite{krizhevsky_nips2012_alexnet} to cutout \cite{devries2017improved} and autoAugment \cite{cubuk2019autoaugment}. Empirical experience however shows that too strong augmentations (i.e., hard positives) are sometimes counterproductive  in the supervised setting. In this section, we explore \textit{why unsupervised contrastive instance learning can benefit from hard positive samples}.
 

\begin{wraptable}{r}{0.35\textwidth}
	\vspace*{-0.020\textwidth}
	\caption{Investigation on different data augmentation settings' effect on MoCov2 \cite{chen_arxiv2020_mocov2} versus supervised training. The setting of color jittering strength follows SimCLR \cite{chen_icml2020_simclr}.}
	\begin{subtable}{\linewidth}
		\tiny
		\begin{tabular}{c|ccc}
			Method & baseline & medium &  extreme  \\
			& 20\%/100\% & 20\%/50\% &  2\%/10\%  \\
			\hline
			\hline 
			MoCo & $64.4$ & $63.7$ &$47.1$ \\
			v2&     & $-0.7$ &$-17.3$ \\
			\hline 
			Super- & $75.5$ & $74.8$ &$52.0$ \\
			vised&     & $-0.7$ &$-23.5$ \\
		\end{tabular}
		\caption{Rand cropping strength (min/max size)}
		\label{tab:data_aug1}
	\end{subtable}%
\hspace{6ex}
	\begin{subtable}{\linewidth}
	
		\tiny
	
		\begin{tabular}{c|cccc}
			Method & \multicolumn{4}{c}{from weak to strong}  \\
			& 1/8 & 1/4 &  1/2 & 1  \\
			\hline
			\hline 
			MoCo & $63.1$ & $63.9$ &$64.2$ &$64.3$ \\
			v2&     & $+0.8$ &$+1.1$&$+1.2$ \\
			\hline 
			Super- & $75.8$ & $75.7$ &$75.6$ &$74.6$ \\
			vised&     & $-0.1$ &$-0.2$ &$-1.2$ \\
		\end{tabular}
		\caption{Color jittering strength}
		\label{tab:data_aug2}
	\end{subtable} 
\vspace*{-0.050\textwidth}
\end{wraptable}

%
%

In contrastive learning without class labels, each image becomes its own class and as a result, there are no clear semantic class boundaries like they exist in supervised training. 
%
%
In order to illustrate how this hurts supervised learning, we provide a t-SNE visualization \citep{Maaten_JMLR2008_tsne} in Figure~\ref{fig:tsne} to show how the class boundaries in ImageNet break down when strong augmentations are applied.
We term this phenomenon as \textit{semantic label shift problem}.
However, during instance discrimination, strong augmentation turn samples into hard positive samples that are recently found to be quite helpful in discriminating instance from instance~\cite{khosla2020supervised}.
%

%
Quantitatively, from Table~\ref{tab:data_aug1}, we can see that as the cropping augmentation becomes more ``extreme'', MoCov2 performance  suffers less than its supervised counterparts.
The reason why performance of MoCov2 also drops might be MoCov2 is learning occlusion invariant features, a view that is corroborated by \citep{purushwalkam_arxiv2020_demystify}. Such occlusion invariant learning process is hindered by aggressive cropping augmentation.
In Table~\ref{tab:data_aug2}, we show that stronger color jittering benefits MoCov2 but hurts supervised learning which is consistent with the findings from SimCLR~\citep{chen_icml2020_simclr}.


%

%


\section{Negative Samples}
\label{sec:negative}

Computing contrastive loss requires sampling negative pairs to avoid learning collapsed representations.
Recently, \cite{wu_cvpr2018_instdisc,he_cvpr2020_moco,chen_icml2020_simclr} have empirically shown that using a large number of negative samples is beneficial to learn good features in contrastive instance learning. 
However, one needs to design sophisticated mechanisms to store the negative examples and ways to update them.
So we ask, \textit{is it possible to use less negative examples during contrastive loss computation without performance degradation}?

\begin{figure}
	\centering
	\begin{minipage}{.45\textwidth}
		\centering
		\includegraphics[width=\linewidth]{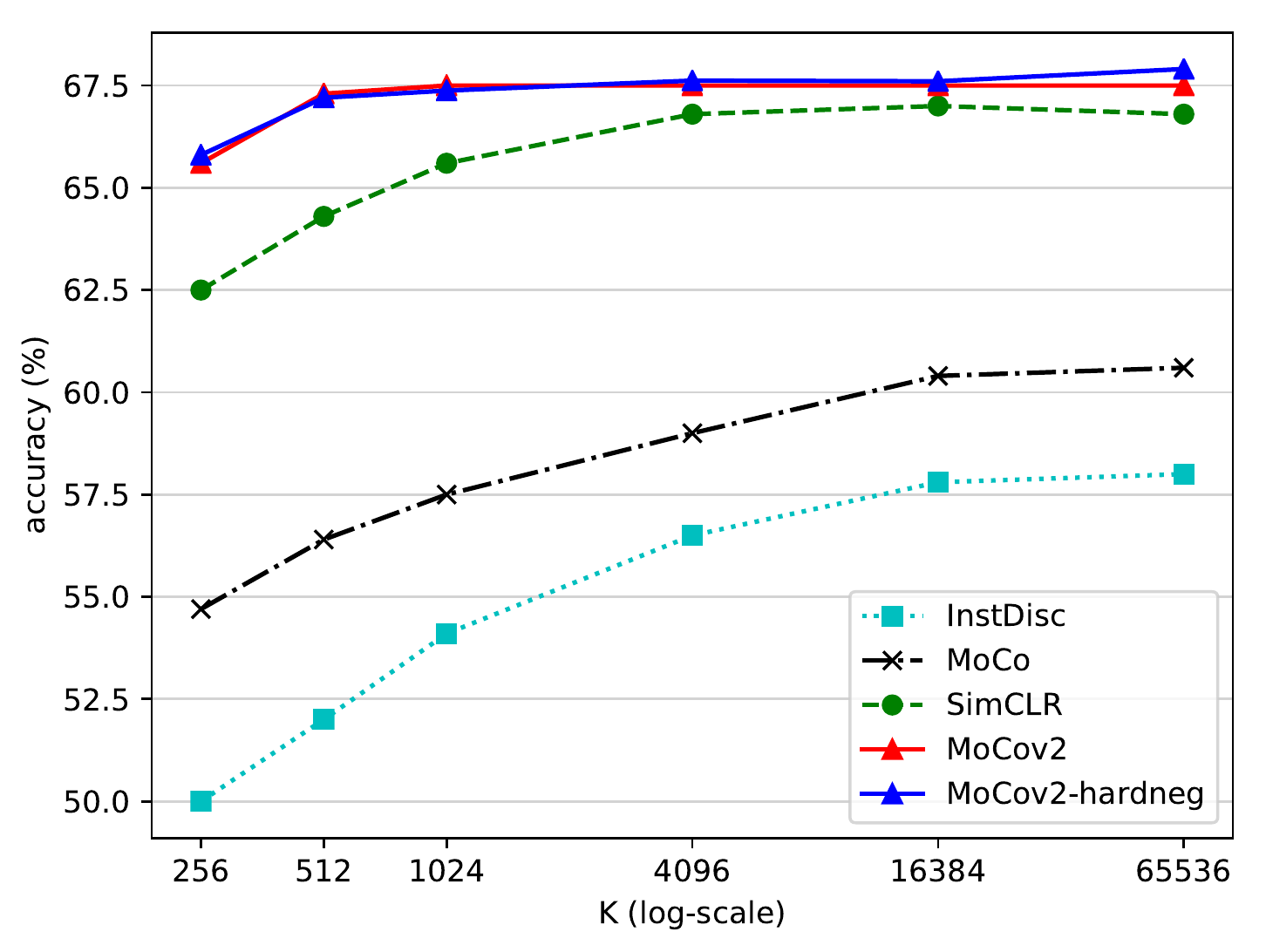}
		\captionof{figure}{Comparison of different algorithms with varying number of negative samples. $K$ denotes the number of negative samples. We show that MoCov2 performs the same ranging from $K=512$ to $K=65536$. }
		\label{fig:negative}
	\end{minipage}%
    \hspace{6ex}
	\begin{minipage}{.45\textwidth}
		\centering
		\includegraphics[width=\linewidth]{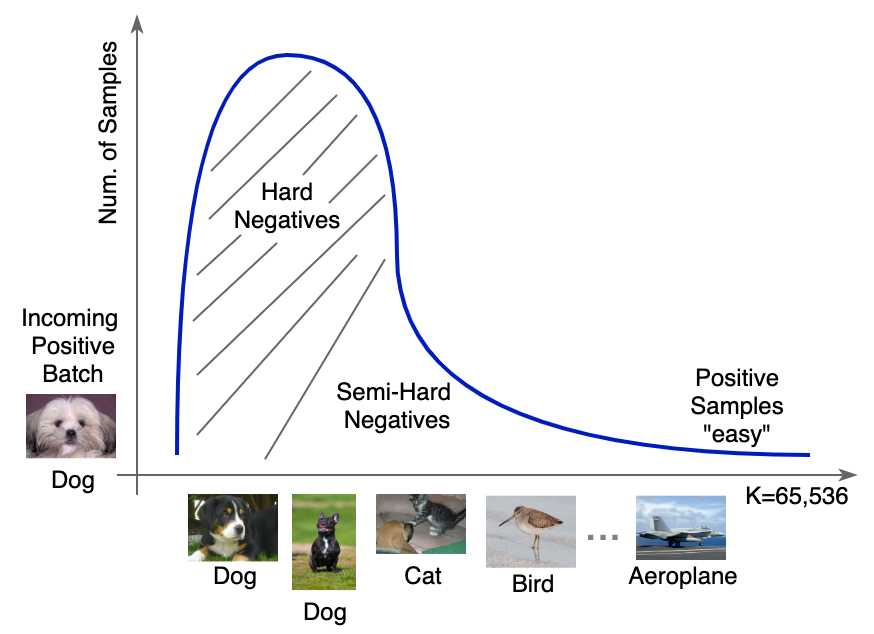}
		\captionof{figure}{Illustration of how we perform dynamic hard-negative sampling on MoCov2 during training time for each mini-batch. We sort the entire dynamic queue based on similarity with incoming positive samples. We use positive skew-normal distribution to select more of hard and semi-hard samples and less of ``easy'' positive samples.  }
		\label{fig:hardneg}
	\end{minipage}
\vspace{-2ex}
\end{figure}

%

\textbf{Quantity of Negative Samples:} In order to answer the question, we first run experiments on recent contrastive instance learning approaches using different number of negative examples. 
We choose InstDisc \citep{wu_cvpr2018_instdisc}, MoCo \citep{he_cvpr2020_moco}, SimCLR \citep{chen_icml2020_simclr} and MoCov2 \citep{chen_arxiv2020_mocov2} as illustrating methods. 
Although they use negative examples in different manners, e.g., memory bank in InstDisc, queue in MoCo and MoCov2, and large mini-batch in SimCLR, they all use large number of negative examples.
We adopt their official released code and follow the same hyper-parameter setting. 

We find, that the performance of InstDisc and MoCo drops when number of negatives decreases unlike for SimCLR and MoCov2 where the impact is less, consistent with \citep{wu_cvpr2018_instdisc,he_cvpr2020_moco,chen_icml2020_simclr}.
In fact, MoCov2 performs consistently with respect to the number of negatives and its accuracy on ImageNet stays the same ranging from $K=512$ to $K=65536$, where $K$ denotes the number of negative samples. 
Hence, the \textit{quantity} of negative samples during contrastive loss computation has little impact on final linear probe performance.
We also find that AP on PASCAL VOC object detection task does not regress due to lower negatives, indicating the quality of the learned features remain the same no matter how many negatives were used ($512$ or $65536$) during contrastive instance pretraining.

\textbf{Quality of Negative Samples}: What if we alter the \textit{quality} of negative samples by performing dynamic hard-negative mining in the MoCov2 queue? Specifically, for a positive batch of images $B$ and an existing dynamic queue $Q$, we sort $Q$ for each batch based on cosine-similarity between the query-branch feature $q$ and the latent features in $Q$. 
We mine hard and semi-hard negative exemplars from dynamic $Q$ using a skew-norm distribution. 
Note, we over-sample in order to keep the overall queue size same as MoCov2.
The intuition for this experiment can be seen in Figure \ref{fig:hardneg}.
Contrasting among these hard-negative samples intuitively should improve the learned representation.
However, as shown in Figure~\ref{fig:negative} (blue line), in spite of over-sampling hard-negative samples, performance of MoCov2 does not change for various sizes of $K$. 


To summarize, in spite of \textit{quantity} and \textit{quality} of negative samples in queue $Q$, MoCov2 performance appears to be stable. We hypothese that the good practices introduced in MoCov2 (MLP head, GaussianBlur augmentations, cosine learning rate scheduling) are responsible for this effect. Thus, we conduct ablation studies to find out the which good practice(s) helps in eliminating the need for large number of negatives. 

\begin{wraptable}{r}{0.5\textwidth}
\vspace{-2ex}
	\footnotesize
	\caption{Investigation of good practices' impact on the number of negative examples (K) being used in computing contrastive loss (wo: without, w: with).}
	\label{tab:negatives}
		\scalebox{0.8} {
		\begin{tabular}{lc|c}
			\hline
			Method         & K=512 & K=65536 \\
			MoCov2    &  $67.3$   &  $67.5$  \\
			(a) wo MLP projection head    & $61.7$  &  $63.6$ \\
			(b) Data Aug: wo GaussianBlur   & $65.5$   &  $66.4$\\
			(c) wo cosine learning schedule & $67.2$  &  $67.3$ \\
			(d) w smaller mom. (0.5) in mom. encoder & $59.8$  &  $64.5$ \\
		\end{tabular} 
	}

\end{wraptable}

\textbf{Good Practices}: From Table~\ref{tab:negatives}, we can see that the MLP head and momentum encoder have the biggest impact (row a and d). 
Without these two techniques, the performance quickly drops as the number of negative samples decreases.
This also explains why SimCLR is robust to the number of negatives with $K=4096$ and $K=65536$, however performance quickly degrades with $K=256$ due to the lack of such mechanisms.
We hope our empirical evidence can provide insights and better understanding of recent progress, as well as advance future development of self-supervised representation learning.

\bibliography{neurips_2020}
\bibliographystyle{nips}
\end{document}